\documentclass[10pt,twocolumn,letterpaper]{article}

\usepackage{cvpr}              

\usepackage{graphicx}
\usepackage{amsmath}
\usepackage{amssymb}
\usepackage{booktabs}
\usepackage{xspace}
\usepackage{pifont}
\newcommand{\MATLAB}{\textsc{Matlab}\xspace}

\usepackage{breqn}
\usepackage{amsmath}

\DeclareMathOperator*{\argmin}{arg\,min}

\usepackage{units}
\usepackage{babel}
\usepackage{mathtools}
\usepackage{stmaryrd}

\usepackage[accsupp]{axessibility}

\usepackage[pagebackref,breaklinks,colorlinks]{hyperref}

\usepackage[capitalize]{cleveref}
\crefname{section}{Sec.}{Secs.}
\Crefname{section}{Section}{Sections}
\Crefname{table}{Table}{Tables}
\crefname{table}{Tab.}{Tabs.}

\def\cvprPaperID{9} 
\def\confName{CVPR}
\def\confYear{2023}

\begin{document}

\title{Human Spine Motion Capture using Perforated Kinesiology Tape}

 \author{Hendrik Hachmann \quad \quad \quad \quad Bodo Rosenhahn\\
  Institute for Information Processing (tnt) / L3S, Leibniz University Hannover\\
 {\tt\small \{hachmann, rosenhahn\}@tnt.uni-hannover.de}
 }

\twocolumn[{%
\renewcommand\twocolumn[1][]{#1}%
\maketitle
\begin{center}
    \centering
    \captionsetup{type=figure}
    \includegraphics[width=.99\textwidth]{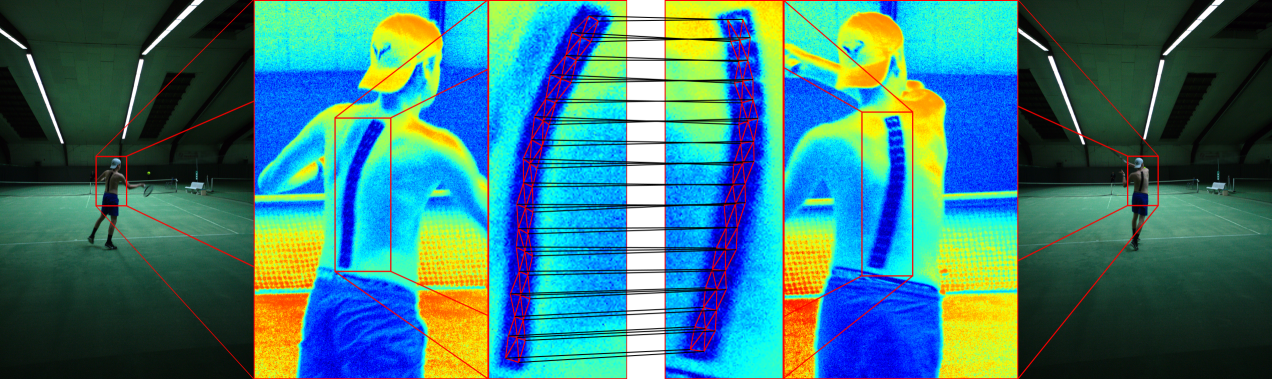}
    \captionof{figure}{Spine tracking during a forehand in Tennis. Perforated kinesiology tape on the back of the player is used as a marker. The arrangement of its dots is known a priori and exploited for matching in a multi view 3D tracking system.}
    \label{burki}
\end{center}%
}]

\begin{abstract}
In this work, we present a marker-based multi-view spine tracking method that is specifically adjusted to the requirements for movements in sports. A maximal focus is on the accurate detection of markers and fast usage of the system. For this task, we take advantage of the prior knowledge of the arrangement of dots in perforated kinesiology tape. We detect the tape and its dots using a Mask R-CNN and a blob detector. Here, we can focus on detection only while skipping any image-based feature encoding or matching. We conduct a reasoning in 3D by a linear program and Markov random fields, in which the structure of the kinesiology tape is modeled and the shape of the spine is optimized. In comparison to state-of-the-art systems, we demonstrate that our system achieves high precision and marker density, is robust against occlusions, and capable of capturing fast movements.
\end{abstract}

\section{Introduction}
\label{sec:introduction}
Human motion capture is a long-studied field that is of particular interest in application fields like health, feature film production, and also increasingly in sports. It allows to capture the movements and techniques of athletes and uncovers specific traits, characteristics, that may give them an advantage but also problems in the performance that might raise health problems. One motivation is straightforward: If the motion of the professional athlete of your choice is captured with sufficient precision, a motion capture system may tell you how to improve yourself to come close to your idol. This is one of several reasons for the widespread publicity in news and media accompanying the 2018 study of Schepers \etal \cite{Schepers:2018}, when they captured Tennis players at Wimbledon using an Xsens MVN Link suit. Their tracking system can be further used for extensive bio-mechanical analyses, i.e. that investigate how forces propagate through the body of an athlete in a musculoskeletal model as OpenSIM \cite{Delp:2007}. This may be used to design person specific training plans to achieve damage precaution and consequently, avoid injuries.

Back pain has become a major disease of civilization that affects large parts of the population. Some causes are widely known, such as lack of exercise or poor posture, e.g. when sitting at a desk at the workplace. But lack of back mobility and incorrect technique in sports might also lead to back pain. In a spinal kinematics study by Campbell \etal \cite{Campbell:2014} with the 2014 Tennis Australia National Squad the authors found a connection between lower back pain and lumbar flexibility during serve kinematics.

The vertebral column consists of 24 vertebrae and has an incredible range of motion \cite{Bogduk:2005} with each two vertebrae capable of lateral flexion, rotations, flexion, and extension to each other. State-of-the-art methods unfortunately do not come close to capture these degrees of freedom. The aforementioned Xsens MVN Link suit consists of 17 inertial measurement units (IMUs), measuring local orientations from which the pose is derived. Considering the spine, only 4 IMUs contribute to its pose estimation: one on each shoulder, one on the chest, and one on the pelvis. As will be discussed in Section \ref{sec:sota}, this is  not sufficient. To allow capturing the high degree of freedom of the spine, the number of measurement points needs to be increased.
\begin{figure}[t]
\begin{center}
 \includegraphics[width=0.99\linewidth]{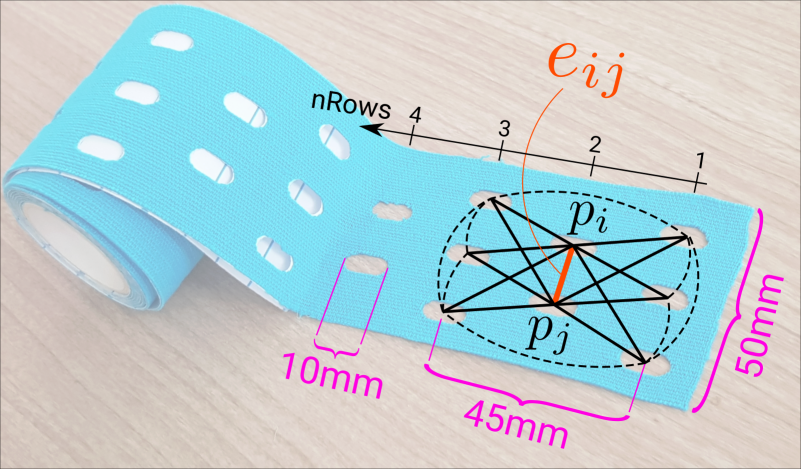}
 \end{center}
 \caption{(background): A roll of kinesiology tape (foreground): Spacial dimensions as the mutual distance are exploited as 3D a priori knowledge of the structure of the tape. (graph): The two points $p_{i}$ and $p_{j}$ define the edge $e_{ij}$. All of the black edges are defined as the neighboring edges of $e_{ij}$ and denoted as $\Omega_{e}$.}
 \label{tape}
\end{figure}

Kinesiology tapes are widely used in all sports mainly for muscle recovery and injury prevention of athletes. Punched or perforated kinesiology tape, as can be seen in the background of Fig. \ref{tape}, is used because of greater stretch capacity and breath-ability. In this work, we divert this kind of kinesiology tape from its intended use and instead apply it in a maker-based motion capture system. As can be seen in Fig. \ref{burki}, the tape is fitted to the back of an athlete and serves as an easy-to-detect pattern with the dots of the tape used as markers in a multi-view camera system. 

The key contributions of this paper are:
\begin{itemize}
 \item The introduction of highly suitable marker for tracking in sports, that can be fast and easily attached and do not restrict movements
 \item An adopted motion capture framework, that uses structural a priori knowledge in 3D instead of image-based feature encoding and matching. 
 \item A high precision tracker that robustly tracks dense markers and is capable of handling occlusions
\end{itemize}
\section{Related Works}
\label{sec:sota}
Accurately tracking the human spine in motion is still an open problem in computer vision. While today human motion capture is often solved marker-less using for instance a combination of key point networks, such as the Convolutional Pose Machines by Wei \etal \cite{Wei:2016} and a 2D-to-3D pose-lifting network as Bogo \etal \cite{Bogo:2016}, such a trend cannot be observed for human spine capturing. This is not surprising, since the back has few cues in its local environment that can be used to infer internal motion. To the best of our knowledge, there is currently no  dedicated marker-less spine motion capture system. A major limitation of general marker-less motion capture systems is that applied 3D skeletons typically consist of 1 to 3 joints for the complete spine, which cannot accurately represent its various contortions.

However, there are a number of methods to infer the motion of the spine using additional technical tools. The low-cost device of Kam \etal \cite{Kam:2017} infers the total curvature of the spine by the outcoupling of light of an optical fiber attached to the back, but it cannot give a precise 3D path. 

Similar to the aforementioned Xsens MVN suit by Schepers \etal \cite{Schepers:2018}, the specialized inertial tracking device by Hajibozorgi and Arjmand \cite{Hajibozorgi:2016} captures the movement of the spine via a series of four IMUs. Voinea \etal \cite{Voinea:2017} use 5 IMUs and a 7th-grade polynomial is fitted to approximate the curvature of the spine. Yet, IMU-based systems suffer from drifting in space, since a two times integration leads to a quadratic error propagation with time.  

With extensive hardware input, Dynamic FAUST by Bogo \etal \cite{Bogo:2017} is able to build a high-precision motion and shape capture system. It uses 22 stereo cameras and 34 speckle projectors to capture 3D body shapes. In addition, body painted features are captured by 22 RGB cameras. The high-frequency texture is mapped onto the body scan and an optical flow-based texture registration provides dense matches. The occurring per-vertex 3D displacements can be used to optimize the body shape and track individual vertices over time. This 4D dynamic scanner creates sequences of 3D body shapes with consistent vertex IDs. Its localization precision, which is closer than 1 mm, is due to the massive amount of data that provides texture matches of images shot from very similar viewpoints.

The retro-reflective marker-based optical motion capture (OMC) system by Vicon \cite{Vicon} is often seen as gold-standard. The study by Merriaux \etal \cite{Merriaux:2017} reports a mean 3D position error of \unit[1.5]{mm} while Rast \etal \cite{Rast:2016} reports a range of 0.1 to \unit[5.3]{mm}, varying with the size of the captured volume. In the specific case of 3D trunk movement capturing with 10 markers along the spine and 12 further on other trunk landmarks, Rast \etal \cite{Rast:2016} compare sources of errors, as the instrumental error of Vicon, different landmark protocols for the markers and soft tissue artifacts. However, the study suffers from limitations, as experiments of the range of motion task for flexion and extension of the spine could not be recorded since the visibility of the markers was insufficient. A likely cause is the minimal number of two cameras needed to track. With the suggested default number of 3 cameras \cite{Vicon} in a 360-degree studio, this threshold might not be reached in a lot of frames, so that trajectories are lost.

The most similar approach to ours is a retro-reflective mesh suit by Hiroaki \etal \cite{Hiroaki:2005}, in which stripes are arranged as a grid and intersections serve as markers. The stripes have a distance of about \unit[30]{mm} from each other, resulting in a high marker density of \unit[0.16]{markers/cm$^2$}. Similar to our argumentation, the authors suggest that a high marker density can only be achieved if connectivity information is present, which they extract directly from the image. Stripe lines are traced and a 2D mesh is extracted, which is fused in 3D with information from other views. This workflow is fundamentally different from ours, as we use no active light setup and enable a higher marker density by explicitly omitting to find 2D connectivity information or any other 2D feature encoding. Instead in 2D we only detect positions and solve the connectivity problem in 3D exploiting a priori known structural information. 
\section{Method}
\label{sec:method}
Our marker-based tracker uses perforated kinesiology tape with a width of \unit[5]{cm} and dots having a diameter of roughly \unit[1]{cm} (see Fig. \ref{tape}). Attached to the back it does not limit the movements of an athlete nor will it fall off in any circumstance. It has a unique pattern of alternating rows with two or three dots next to each other. As can be seen in the teaser in Fig. \ref{burki}, the large difference in brightness helps in cases of underexposed photos thereby allowing for short exposure times of the camera. By their nature, circles are stable 2D markers and a robust center point extraction is possible even in very noisy or (motion) blurred images. The kinesiology tape provides a high marker density, with 22 dots on a \unit[20$\times$5]{cm} stripe resulting in a marker density of \unit[0.22]{markers/cm$^2$}.

Our hardware setup consists of 6 global shutter RGB cameras (FLIR ORX-10g-51S5-C) set to a resolution of 2448$\times$1600 and a frame rate of \unit[100]{fps}. The exposure time of the camera is set to \unit[1]{ms}, which is rather short and we consequently observe very little motion blur when testing with fast stripe movements.  As can be seen in Fig. \ref{studio} the cameras are circularly arranged covering 180 degrees of the athlete. The cameras are synchronized and calibrated to each other and to a Vicon infrared marker reference system (Vicon Motion Systems, Oxford, UK \cite{Vicon}). The Vicon system uses 8$\times$T010 cameras and Nexus software in version 1.8.5. 
\newcommand{\mywidth}{0.23}
\begin{figure}[t]
\begin{center}
    \includegraphics[width=\mywidth\textwidth]{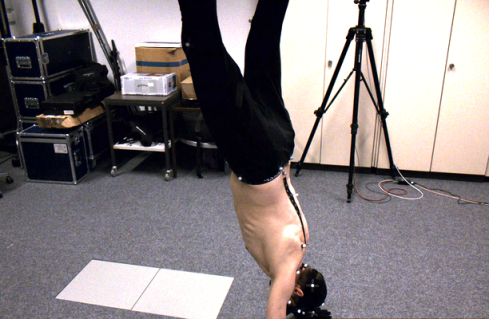}
    \includegraphics[width=\mywidth\textwidth]{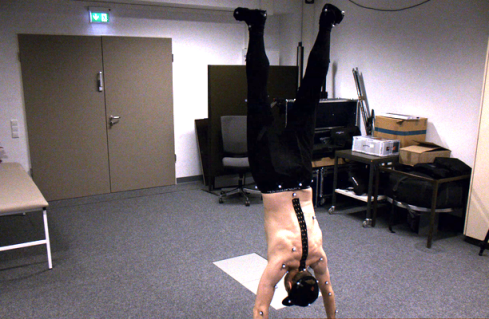}\\
    \includegraphics[width=\mywidth\textwidth]{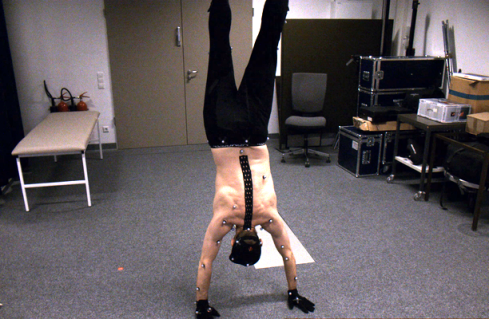}
    \includegraphics[width=\mywidth\textwidth]{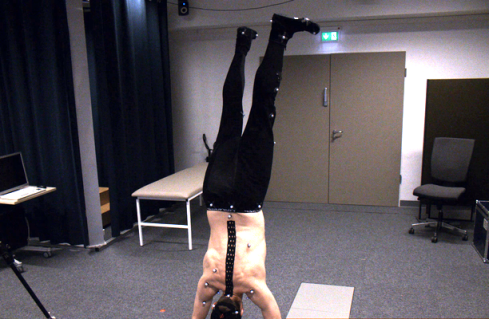}\\
    \includegraphics[width=\mywidth\textwidth]{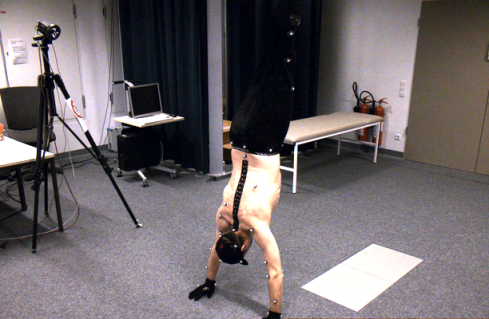}
    \includegraphics[width=\mywidth\textwidth]{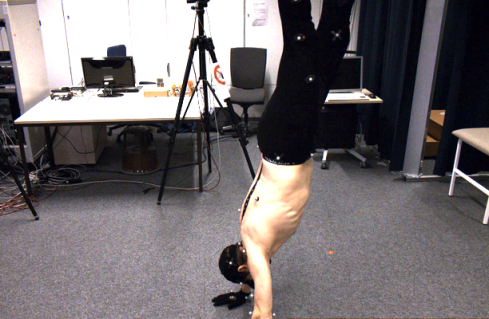}
 \end{center}
 \caption{As motivated in \cite{RosSch2008}, an athlete is performing a handstand in a studio setup with 6 cameras in an u-shape arrangement covering 180 degrees. The kinesiology tape attached to the back can be seen in 4 views. In addition a reference Vicon \cite{Vicon} camera setup tracks infrared makers attached to the rest of the body.}
 \label{studio}
\end{figure}

\begin{figure*}[t]
\begin{center}
    \centering
    \captionsetup{type=figure}
    \includegraphics[width=0.99\linewidth]{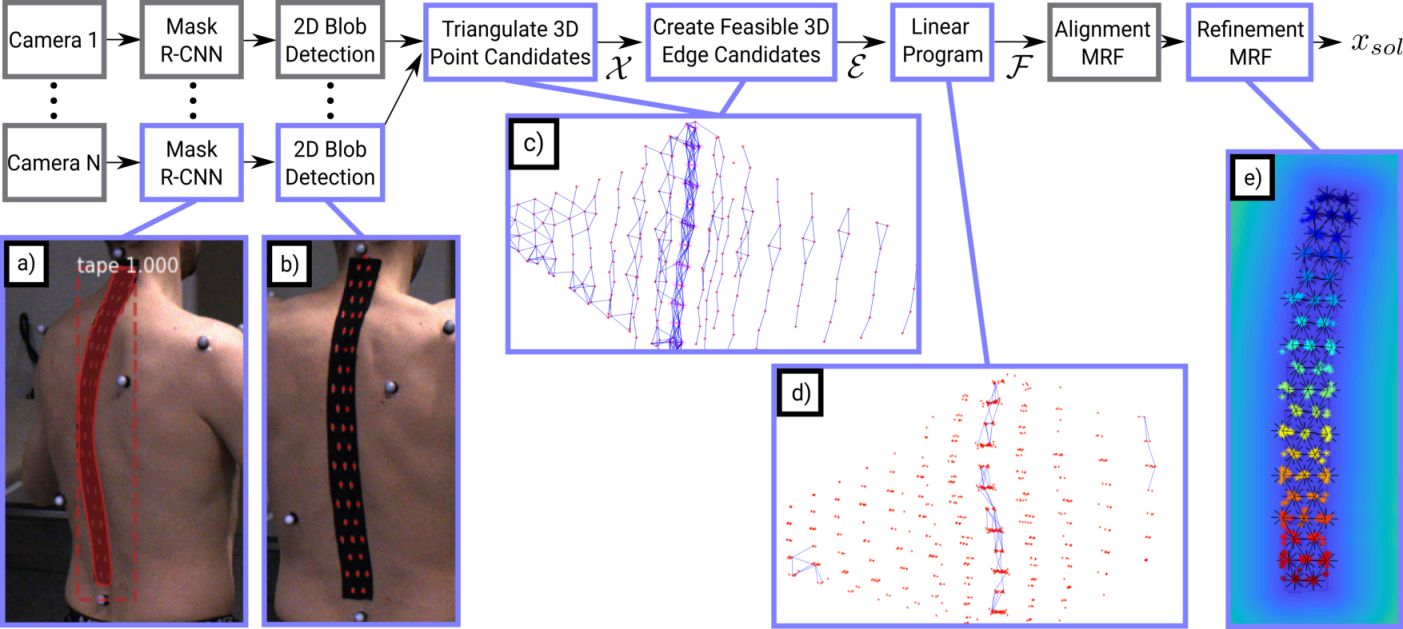}
    \captionof{figure}{The proposed tracking pipeline: a) a Mask R-CNN \cite{He:2017} detects the kinesiology tape in $N$ camera images. The corresponding mask is used as a region of interest (ROI) in which b) a blob detector finds the dots of the tape. c) dot marker points of all cameras are mutually triangulated which gives the 3D point candidate cloud $\mathcal{X}$, in which edges of appropriate lengths are found $\mathcal{E}$. d) a linear program selects most likely edges and gives a feasible edge set $\mathcal{F}$. A 3D-based alignment Markov Random Field (MRF) model is iteratively registered to $\mathcal{F}$ given the a priori known structure of the dots inside of the tape. e) finally a second MRF optimized by particle belief propagation refines the model to all camera views and outputs the estimated 3D marker coordinates $x_{sol}$.}
    \label{blockdiagram}
\end{center}
\end{figure*}

An overview of our signal processing pipeline is illustrated as a block diagram in Fig. \ref{blockdiagram}. As a first step, the stripe and its dots are detected in each camera individually. A Mask R-CNN as proposed by He \etal \cite{He:2017} is used as a stripe detector. The network is pre-trained on the MS COCO \cite{Lin:2014} dataset and we refine the network on a dataset consisting of 1988 training samples and 631 validation samples. Each sample is a tuple of an RGB image acquired in our studio and the corresponding mask. A detected mask $m$ (see Fig. \ref{blockdiagram} a)), is used as a region of interest (ROI) for a blob detector (Fig. \ref{blockdiagram} b)), detecting the dots in the stripe. For our purpose, the blob analysis of the \MATLAB Computer Vision Toolbox \cite{Matlab:2019} is sufficient to provide blob center points $b \in \mathcal{B}$.

The points $b$ are mutually triangulated using blobs of another view and the resulting 3D point is considered as a 3D candidate point $p \in \mathcal{P}$, if its back projection error is below a threshold $th_{1}$. The resulting point cloud $\mathcal{P}$ consists of points that can be seen by two cameras. Each point $p_i$ has a set of attributes: the identifiers $a_{i}$ of cameras used for triangulation, the blobs $b_{i}$ used for triangulation and a normal $n_{i}$ derived as the negative normalized view directions of participating cameras. Next, two points $p_1, p_2 \in \mathcal{P}$ are merged into a new point $p_3 = (p_1+p_2)/2$, if the Euclidean distance towards each other is below threshold $th_{2}$ and if the back projection error of the merged point in all views $a_{1}$ and $a_{2}$ is below the threshold $th_{1}$. If $p_1$ and $p_2$ are merged, they are removed from $\mathcal{P}$ and $p_3$ is added to $\mathcal{P}$ with the attribute $a_{3} = a_{1} \cup a_{2}$ and $b_{3}$, $n_{3}$ respectively. This process is repeated as long as points can be merged. As the next step, the point cloud $\mathcal{P}$ is transformed into an edge cloud $\mathcal{E}$ with $\{e \in \mathcal{E} \mid l_{e} > th_{3} \text{ and } l_{e} < th_{4}\}$, where $l_{e}$ is the length of an edge connecting two points in $\mathcal{P}$. The thresholds $th_{3}$ and $th_{4}$ are the lower and upper boundaries of the a priori known distances of dots (see Fig. \ref{tape}). In this way, we allow variations in distance between the dots of the tape if the skin at the back is stretched or contracted. The described sets can be seen in Fig. \ref{blockdiagram} c) with $p \in \mathcal{P}$ printed in Red and $e \in \mathcal{E}$ in Blue.

\subsection{Linear Program}
The task of the linear program is to select a subset from the set of edges $\mathcal{E}$ as an approximation of the stripe (see Fig. \ref{blockdiagram} c) and d)). For this task, we define a graph $\mathcal{G} = (\mathcal{P}, \mathcal{E}, \mathcal{C})$ on the stripe, where $\mathcal{C}$ is a set of costs associated to the points. As can be seen in Fig. \ref{tape}, points $p \in \mathcal{V}$ are the dots of the stripe, and $e \in \mathcal{E}$ are edges connecting points. Given, that each stripe starts and ends with a two dots row, the number of target edges can be calculated by
\begin{equation}
    n_{e} = \lfloor n_{row}/2\rfloor \cdot 2 + \lceil n_{row}/2\rceil + (n_{row}-1) \cdot 6,
\end{equation}
with $\lfloor .\rfloor$ and $\lceil .\rceil$ the floor and ceiling function.
We assign individual edge costs $c_{e} \in \mathcal{C}$ as the inverse of the number of neighboring edges $\Omega_{e}$ connected to the current edge. In Fig. \ref{tape} the black edges of the graph are neighboring edges of the red edge $e_{ij}$, which is defined by the two points $p_i$ and $p_j$. Consequently, $\Omega_{e}$ is defined as all edges connected to $p_i$ or $p_j$ except $e_{ij}$.

The linear program solves an edge selection task, with selection hypothesis $\mathcal{H}$ and indicator variables $x_{e}$, which take value 1 if $\mathcal{H}$ is selected, and 0 otherwise. The stripe detection task is to select hypotheses that minimize the total costs. This can be cast into the following binary optimization problem:
\begin{equation}
 \argmin\limits_{x \in \mathcal{F}} \sum_{e \in \mathcal{E}} c_{e} x_{e}
 \label{eq_argmax}
\end{equation}
where the feasibility set $\mathcal{F}$ is subject to
\begin{equation}
 \sum_{e \in \mathcal{E}} x_{e} \leq n_{e},
  \label{eq_ne}
 \end{equation}
 \begin{equation}
 \forall e \in \mathcal{E}: \sum_{\omega_{e} \in \Omega_{e}} \omega_{e} x_{e} \leq 12,
 \label{eq_12}
\end{equation}
\begin{equation}
  \forall e \in \mathcal{E}: \sum_{\omega_{e} \in \Omega_{e}} \omega_{e} x_{e} \geq 6,
  \label{eq_6}
\end{equation}
The objective function in Eq. \ref{eq_argmax} minimizes the costs and thereby maximizes the number of neighboring edges. Eq. \ref{eq_ne} limits the total number of selected edges to $n_e$. Eq. \ref{eq_12} and Eq. \ref{eq_6} constrain the number of neighboring edges to a range of 6 to 12, which as can be seen in Fig. \ref{tape} reflects the pattern of the dots. 

In addition to Eq. \ref{eq_ne} - \ref{eq_6}, the following condition has to be satisfied: For all edge pairs $e_{ij},e_{kl} \in \mathcal{E}$ that are non-identical ($i \neq k, j \neq l$) and are triangulated using the same blobs in one camera, i.e. $b_{i} = b_{k}, b_{j} = b_{l}$, one indicator $x_{ij}$ or $x_{kl}$ must be zero.

If the binary linear program is infeasible, which it may be in case of occlusions or missing blob detections, we iteratively decrease the number of target edges $n_{e}$ until a feasible solution can be found. The resulting problem can be solved to optimality using BLP solvers like Gurobi \cite{Gurobi:2023}.

\subsection{Alignment Markov Random Fields (MRF1)}
In the following, two Markov random fields MRF1 and MRF2 are sequentially applied (see Fig. \ref{blockdiagram}). In MRF1 a model is optimized to match the edges of the feasibility set $\mathcal{F}$ and in MRF2 the model is further refined to match the camera images. 

The first Markov random field is a set of random variables called nodes ${\bf x} = (x,y,z,d) \in \mathcal{V}$, with $x,y,z$ the 3D location of the node and a local distance $d$. Each node has a neighborhood $\mathcal{N}$, defining connections to other nodes, which consists of the solid and dashed edges drawn in Fig. \ref{tape}. Since we expect a different amount of stretching of the stripe in longitudinal and transverse direction, we distinguish between longitudinal $\mathcal{N}_{1,s}$ and transverse $\mathcal{N}_{2,s}$ edges. 
Transverse edges connect points within a row, while longitudinal edges connect different rows. The random variables ${\bf x}$ are optimized by minimizing the energy of the MRF model:
\begin{equation}
  E({\bf x}) = \sum_{s\in \mathcal{V}} \psi_{s}({\bf x_s}) + 
  \sum_{s\in\mathcal{V}}\sum_{t\in \mathcal{N}_{s}} \psi_{s,t}({\bf x_s}, {\bf x_t}),
  \label{eq:model}
\end{equation}
where $\psi_{s}({\bf x_s})$ is the unary potential and $\psi_{s,t}({\bf x_s},{\bf x_t})$ the binary potential. The unary potential is defined as the shortest distance to a point ${\bf p}$ of the feasibility set $\mathcal{F}$:
\begin{equation}
  \psi_{s}({\bf x_s}) = \min({\bf x_s}, {\bf p}), \forall {\bf p} \in \mathcal{F}
\end{equation}
The minimal energy in Eq. \ref{eq:model} is consequently close to the linear programs solution. The binary potential is defined as
\begin{equation} \label{eq17}
\begin{split}
\psi_{s,t}({\bf x_s},{\bf x_t}) & = \Theta_{1}\left(\left\|{\bf p}_{st}\right\|_{2} - d \right)^{2}  + \Theta_{2}\left\|\langle {{\bf p}_{st}}, {\bf n}_{s,t} \rangle\right\|_{2}^{2}\\
 & + \Theta_{3}\text{exp}(d_{min}-\left\|{\bf p}_{st}\right\|_{2})^{2}\llbracket \left\|{\bf p}_{st}\right\| < d_{min} \rrbracket\\
& + \Theta_{4}(\left\|{\bf p}_{st}\right\|_{2}-d_{target})^{2} \\
\end{split}
\end{equation}
where ${\bf p}_{st} = {\bf p}_{s} - {\bf p}_{t}$ and $\llbracket \cdot \rrbracket$ is the Iverson bracket, which is 1 if its argument is true and zero otherwise. The parameters $\Theta_{1}-\Theta_{4}$ weigh individual terms. In the first term, $\Theta_{1}$ balances the random variable $d$ to match the node distance between ${\bf p}_{s}$ and ${\bf p}_{t}$. The second term pushes the orientation of the connection ${\bf p}_{s}$ to ${\bf p}_{t}$ to be orthogonal to nearby normal ${\bf n_{s,t}} \in \mathcal{N}_f$, where the set $\mathcal{N}_f$ is a smooth field of normals derived from the point attributes ${\bf n_i}$. In $\mathcal{N}_f$ each camera is equally weighted, yet, individual weights for each camera would be beneficial, as could be derived from an advanced clustering of the corresponding cameras as in Dockhorn \etal \cite{DocBra2016}. The third term encourages ${\bf x_s}$ and ${\bf x_t}$ to keep a minimal distance $d_{min}$ from each other and the fourth term constrains the distance ${\bf p}_{st}$ to match the target length $d_{target}$, which are measured tape distances that can be estimated from Fig. \ref{tape}.

The MRF is an iterative algorithm that needs an initialization of a template, which is a complete structure of the stripe including the target number of points with corresponding edges. In order to position the structure in 3D we fit a third degree polynomial to the feasibility set $\mathcal{F}$. Together with the normal field $\mathcal{N}_f$ we can create a tape template along the run of the polynomial. The long connections in Fig. \ref{tape} that span across two rows or columns are included to avoid a folding of the structure of the stripe.

In order to efficiently minimize the MRF energy (Eq. \ref{eq:model}), we approximate the maximum a posteriori (MAP) probability using max-product particle belief propagation as Pacheco \etal \cite{Pacheco:2015} and Hachmann \etal \cite{HacKru2021a}. Particles are estimated using slice sampling as in M\"{u}ller \etal \cite{Mueller:2013}. In order to improve convergence, the slice sampling particle set $P_t = \{{\bf x_t^{(1)}},\cdots,{\bf x_t^{(p)}}\}$ of node $t$ is augmented:
\begin{equation}
  P_{t,aug} = P_t \cup P_{mobility} \cup P_{knn}.
  \label{particleSet}
\end{equation}
The set $P_{mobility}$ adds the position of neighboring nodes which increase stripe movements and thus convergence and $P_{knn}$ adds the k-nearest points from the feasibility set $\mathcal{F}$ as these are likely candidates for a good position. 

\subsection{Refinement Markov Random Field (MRF2)}
The second MRF uses the same energy formulation as Eq. \ref{eq:model} and is initialized by the resulting positions of MRF1. MRF2 is computationally much more demanding, as it tends to need more iterations to convert. That is why we decrease the search space and reduce the random variable ${\bf x}$ to the 3D positions of the nodes. For the same reason, the number of edges is reduced to only include the solid connections in Fig. \ref{tape}. The unary potential of MRF2 is image-based:
\begin{equation}
  \psi_{s}({\bf x_s}) = \sum_{c \in \mathcal{C}} I_{c}(\Phi_{c}({\bf x_{s}}))
\end{equation}
Here, $\Phi_{c}(\cdot)$ is the back projection of 3D points into image coordinates of camera $c \in \mathcal{C}$. The 2D image $I_{c}$, which can be seen in the background of Fig. \ref{blockdiagram} e), contains the inverse image acquired by camera $c$ inside of the Mask R-CNN mask $m$ and is filled with increasing values on the outside of $m$, which are calculated by a distance transformation. The binary potential of MRF2 is similar to Eq. \ref{eq17}:
\begin{equation} \label{eq18}
\begin{split}
\psi_{s,t}({\bf x_s},{\bf x_t}) & = \Theta_{5}(I_{c}(\Phi_{c}({\bf x_s})+I_{c}(\Phi_{c}({\bf x_t})-2I_{c}(\Phi_{c}({\bf x_{st}}))\\ 
 & + \Theta_{6}\left\|\langle {{\bf p}_{st}}, {\bf n}_{s,t} \rangle\right\|_{2}^{2} 
 + \Theta_{7}\left\|\langle {{\bf p}_{st}}, {\alpha}_{s,t} \rangle\right\|_{2}^{2}\\
& + \Theta_{8}(\left\|{\bf p}_{st}\right\|_{2}-d_{target})^{2} \\
\end{split}
\end{equation}
with ${\bf x_{st}}$ the arithmetic mean of ${\bf x_{s}}$ and ${\bf x_{t}}$. Different to Eq. \ref{eq17} the first term calculates a relation between the nodes positions and an intermediate point ${\bf x_{st}}$: Given that the appearance of the stripe dots are bright and regions in between are dark. Furthermore, the third term 
aligns the model to the currently estimated direction $\alpha_{s,t}$, that is either longitudinal or transversal.
The corresponding dot product between the edge ${\bf p}_{st}$ and the identified direction $\alpha_{s,t}$ is penalized. The directions in $\alpha_{s,t}$ are derived from the normal field $\mathcal{N}_f$ and the fitted polynomial. The same particle set $P_{t,aug}$ as in Eq. \ref{particleSet} is used and the back projection $\Phi_{c}(P_{t,aug})$ it is illustrated in the foreground of Fig. \ref{blockdiagram} e) as colorful markers. The parameters as $\Theta_{1},...,\Theta_{8}$ and $d_{min}$ are estimated empirically.

After convergence of MRF2 the final positions of the nodes ${\bf x_{sol}}$ can be fitted to a musculoskeletal model like OpenSIM \cite{Delp:2007} for bio-mechanical analysis. We share software examples of our tracking framework on GitHub\footnote{\url{https://github.com/hendrik-hachmann/spinemocap}}.

\section{Experiments and Results}
\label{sec:results}
We conduct a series of experiments on artificial and studio sequences to prove the effectiveness of our method. While the overall framework remains unchanged, each domain has different empirically estimated parameters. The \textit{Artificial} dataset is the 548 frames long sequence \textit{50009\_hips} of the Dynamic FAUST dataset by Bogo \etal \cite{Bogo:2017}, which shows the largest amount of spine movements among them. The coordinate system suggests that the height of the body shape is 1.75, which we set to 1.75 meters in order to calculate reasonable errors in millimeters.

\begin{figure}[b]
\begin{center}
 \includegraphics[width=0.99\linewidth]{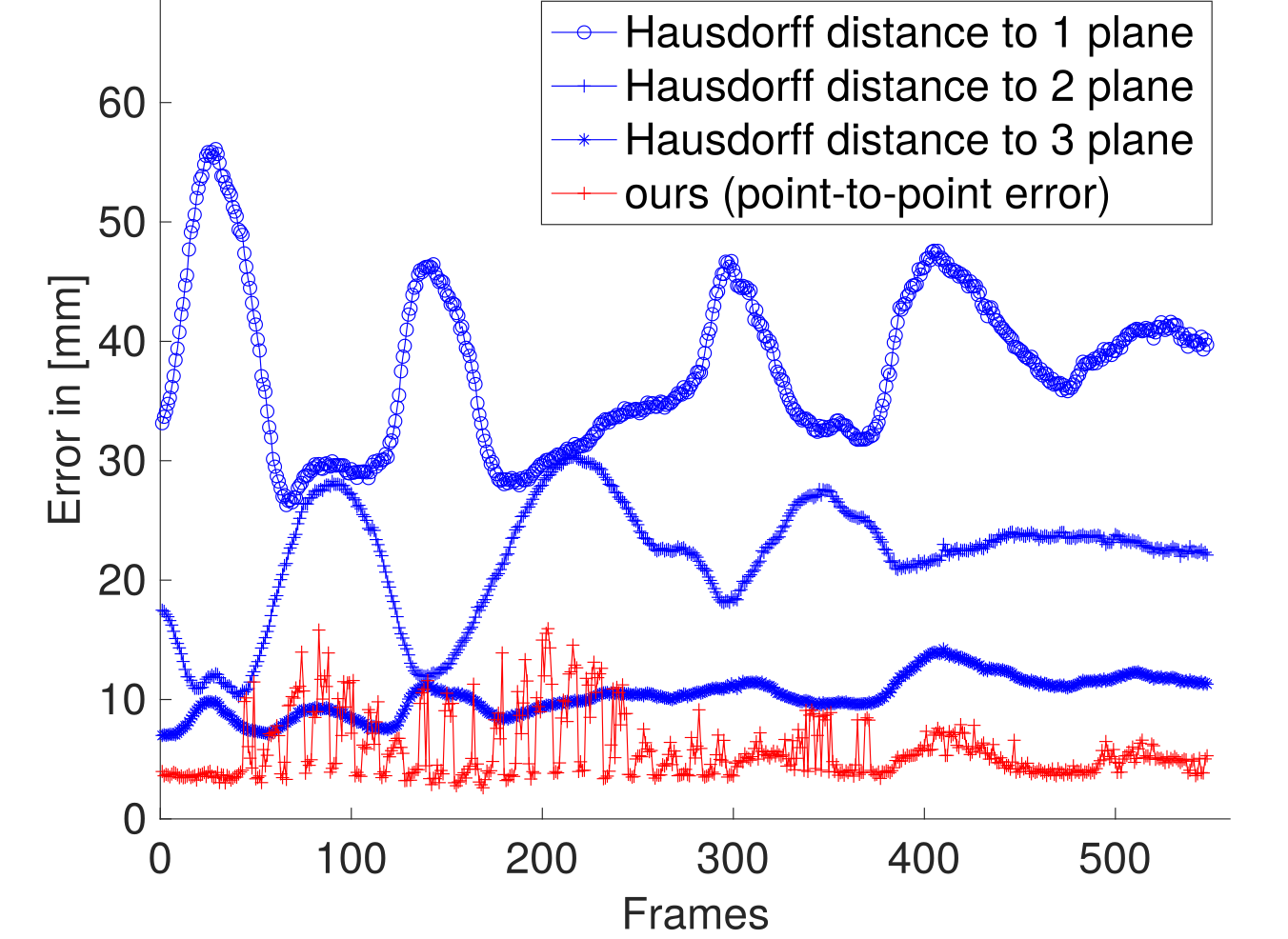}
 \end{center}
 \caption{Comparison of the 1 to 3 planes approximation to the ground truth with our marker point to ground truth point error on the 0 balls artificial sequence. Even though a point to point error is more restrictive than a plane to point error, our method shows that it is more accurate.}
 \label{GraphArtificial}
\end{figure}

\textbf{Experiment 1:} Using the artificial sequence we compare our method to an arbitrary keypoint-based approach that is able to robustly track 4 to 8 keypoints positioned left and right next to the spine. These points are used to fit 1 to 3 planes that approximate the back, which can be seen in Fig. \ref{planes}. The shortest distance of the planes to the tape dot ground truth is measured, which is a one-sided Hausdorff distance. The errors of planes and our approach can be seen in the time series in Fig. \ref{GraphArtificial} and mean values can be found in the top row of Table \ref{tabelle2}.

\newcommand{\myheighttt}{0.64}
\begin{figure}[t]
\begin{center}
 \includegraphics[height=\myheighttt\linewidth]{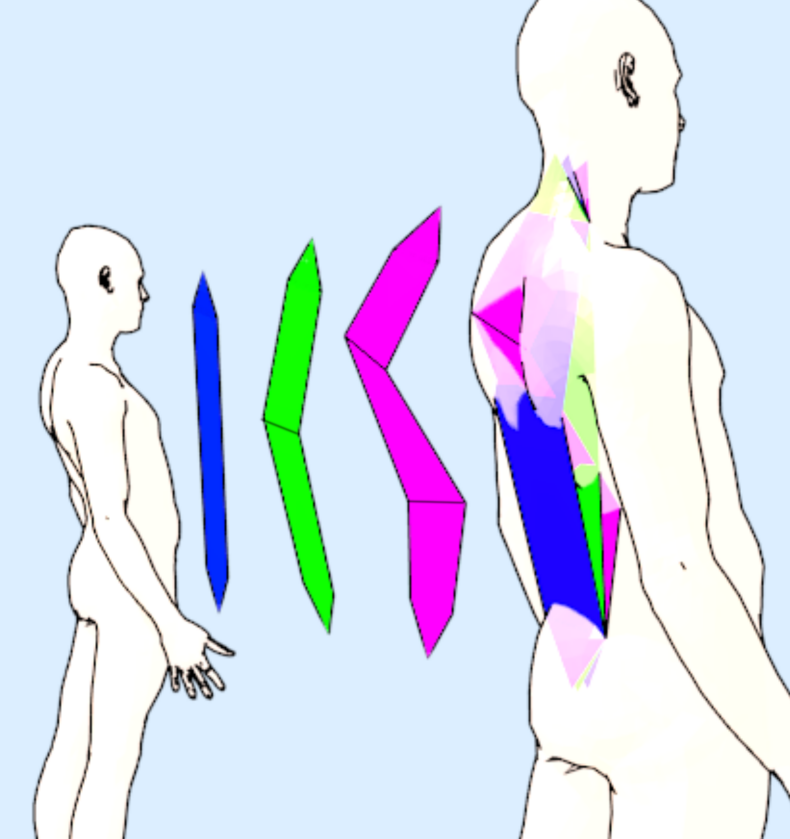}
 \includegraphics[height=\myheighttt\linewidth]{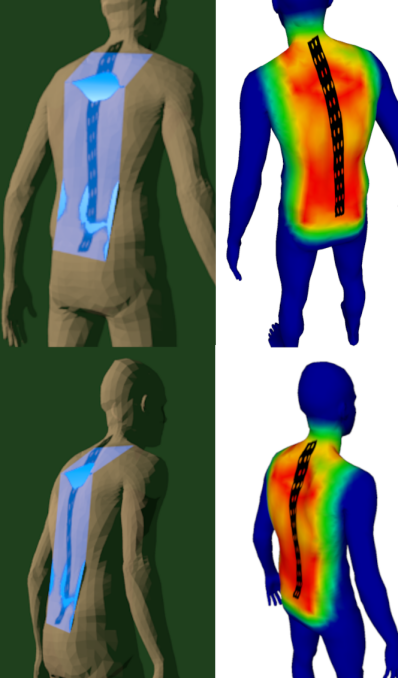}
 \end{center}
 \caption{(left) We perform experiments with one (blue) to three (magenta) planes fitted to markers left and right next to the spine. (center) Qualitative renderings for a 3-plane approximation with the clearly visible plane-to-body shape intersections. (right) Hausdorff distance color encoding with red values being a close approximation to green values indicating a large distance to the planes. This sparse marker-based spine approximation is evaluated on the artificial test sequence and the closest distance of the plane to the 3D ground truth is plotted in the time series in Fig. \ref{GraphArtificial}.}
 \label{planes}
\end{figure}

\begin{figure}[t]
\begin{center}
 \includegraphics[width=0.99\linewidth]{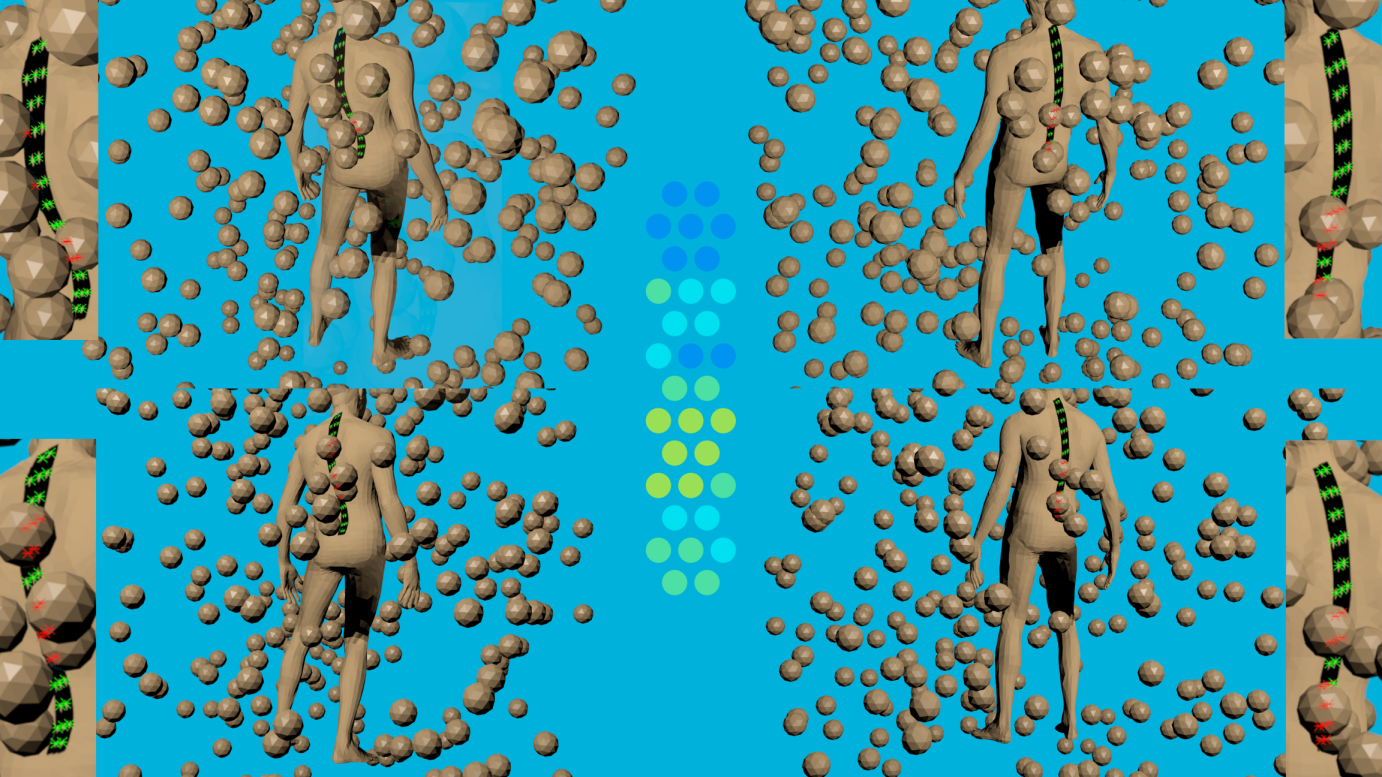}\\
 \includegraphics[width=0.99\linewidth]{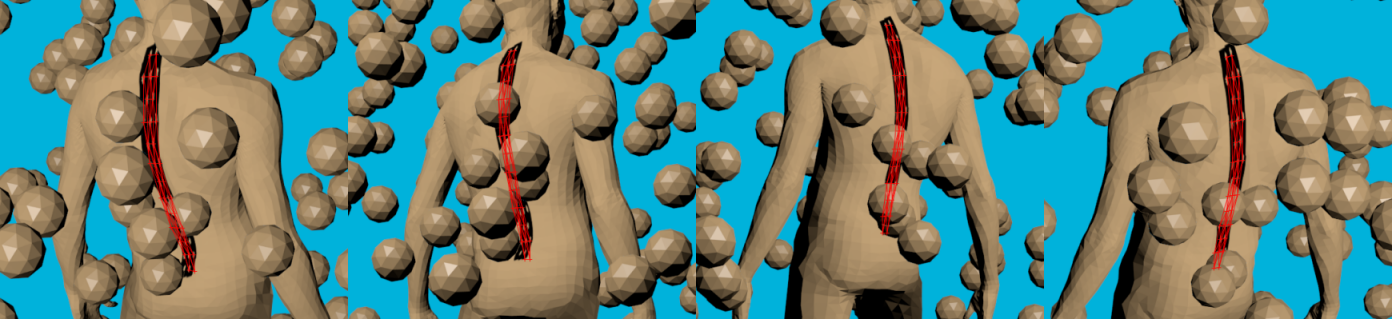} 
 \end{center} 
 \caption{Occlusion experiment: (top rows) 4 views of the \textit{Artificial} sequence with 5000 particle system balls falling from the sky and creating occlusions. Green or red markers on the stripe symbolize whether the particular dot can be seen by the particular view. The schematic in the middle shows the number of cameras that can see the markers with dark Blue meaning that all cameras can see the dot and yellow meaning that only one camera can see the dot. (bottom row) Corresponding spine tracker result. It can be observed that the tracker can estimate the position of stripe markers quite well even though this is a case with a heavy amount of occlusions. While in this case, occlusions are caused by balls, self-occlusions are very common, as in a 180 degrees studio setup the total stripe can never be seen by all cameras at once.}
 \label{occlusionExperiment}
\end{figure}

\textbf{Experiment 2:} We added skin-colored balls falling from the sky to simulate arbitrary occlusions. As illustrated on top of Fig. \ref{occlusionExperiment}, in a rendered scene the red and green marked ground truth positions and occlusions can be easily calculated. The occlusions are mapped to an artificial stripe pattern in the middle showing the number of views that can see a marker. Occlusions typically occur as burst occlusions, meaning that several dots close to each other are occluded at the same time. A qualitative result can be seen at the bottom of Fig. \ref{occlusionExperiment}. Even with the high number of occlusions in this frame, the marker position estimation stays robust. Quantitative results can be found in Table \ref{tabelle1}.

\begin{table}
\begin{center}
\resizebox{\linewidth}{!}{
\centering
\begin{tabular}[h]{c||c|c|c|c|c}
    Dataset & 1-plane & 2-planes & 3-planes & MoSh \cite{Loper:2014} & \textbf{ours} \\
    \hline\hline
    Artificial $\downarrow$ & 37.67 & 21.89 & 10.29 & - & 5.55\\ 
    Studio $\downarrow$  & - & -  & - & 24.75 & 4.12\\
\end{tabular}
}
\end{center}
\caption{Quantitative tracking results: The dataset \textit{Artificial} is a 4D sequence of the Dynamic FAUST dataset \cite{Bogo:2017} with kinesiology tape texture (see Fig. \ref{occlusionExperiment}). The 1-3 plane(s) results are Hausdorff distance errors of the planes illustrated in Fig. \ref{planes} to the ground truth markers. Error-values in [mm] are calculated with the virtual body size set to 1.75 meters. The \textit{Studio} dataset is a studio sequence which can be seen in Fig. \ref{studioresults}. MoSh \cite{Loper:2014} denotes a motion and shape fitting to a sparse Vicon \cite{Vicon} marker trajectory. Individual marker positions are inferred from the texture fitting as illustrated in Fig. \ref{smpl_texture}.}
\label{tabelle2}
\end{table}

\begin{table}
\begin{center}
\resizebox{0.7\linewidth}{!}{
\centering
\begin{tabular}[h]{c||c|c|c}
    Artificial dataset &  0 balls & 5000 balls & 10000 balls \\
    \hline\hline    
    4 cameras  & 17536 & 11197  & 6846 \\
    3 cameras & 0 & 5328  & 6812 \\
    2 cameras & 0 & 825  & 2764 \\
    1 camera & 0 & 83  & 607 \\
    0 cameras & 0 & 103  & 507 \\    
    \hline
    error [mm] $\downarrow$ & 5.55 & 6.17 & 7.89 \\ 
\end{tabular}
}
\end{center}
\caption{Occlusion experiment results corresponding to Fig. \ref{occlusionExperiment}: (top) Visibility of markers in the \textit{Artificial} sequence with 0, 5000, and 10000 balls as occluders. 4 cameras mean no occlusions, as 4 cameras can see the markers. 0 cameras mean that a marker is not visible in any view. (bottom) Marker point-to-ground truth point tracking the error in [mm].}
\label{tabelle1}
\end{table}

\begin{figure}[t]
\begin{center}
 \includegraphics[width=0.99\linewidth]{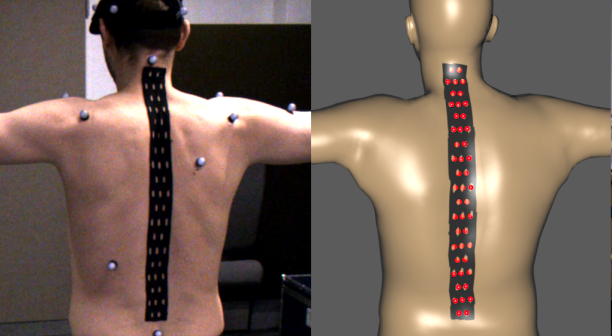}
 \end{center}
 \caption{Vicon marker-based SMPL fitting with stripe texture mapping. (left) T-pose reference frame (right) SMPL fit with texture and red balls as 3D ground truth.}
 \label{smpl_texture}
\end{figure}

\textbf{Experiment 3:} We captured a studio sequence with major spine movement, which we call \textit{Studio} using our RGB cameras and Vicon. For every 50th frame, we create a 3D ground truth by triangulation of manually annotated images of the six cameras. Then we optimize a body shape to the sparse Vicon marker set similar to MoSh by Loper \etal \cite{Loper:2014}. As can be seen in Fig. \ref{smpl_texture}, on a t-pose frame an artificial stripe texture is registered to the ground truth rendered as red balls. As SMPL \cite{Loper:2015} body shapes have consistent vertex IDs along the sequence, from the texture we can calculate a 3D estimate of the tape dots. Ground truth marker-to-marker estimates of this method and our tracker can be found in Table \ref{tabelle2}.

\begin{figure}[t]
\begin{center}
 \includegraphics[width=0.99\linewidth]{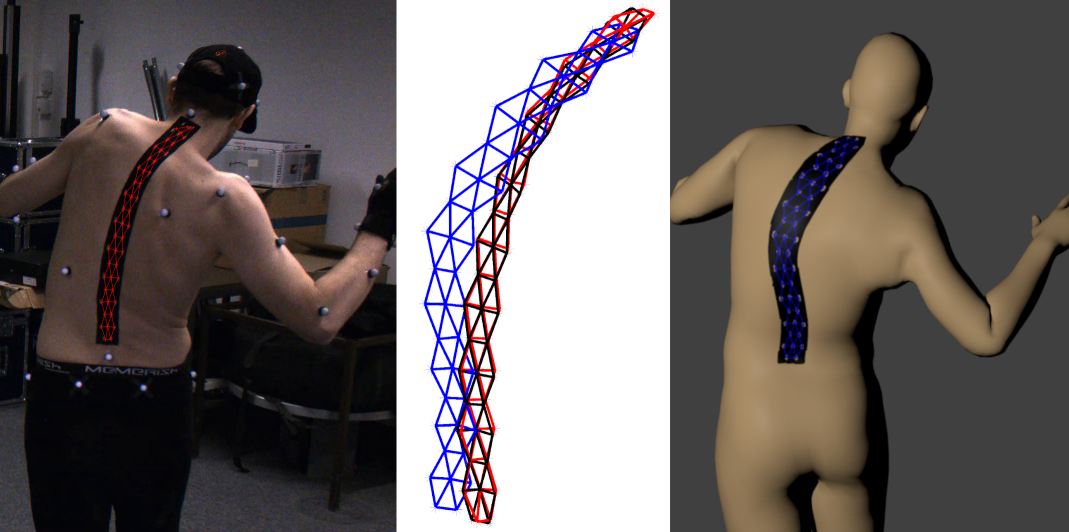}
 \end{center}
 \caption{(left) Studio sequence image with the result of our method (center) Comparison of results of MoSh printed in Blue, our method in Red to the ground truth in Black (right) Vicon marker-based SMPL fitting with tracking result in Blue.}
 \label{comparisonToSMPL}
\end{figure}

\begin{figure}[t]
\begin{center}
 \includegraphics[width=0.92\linewidth]{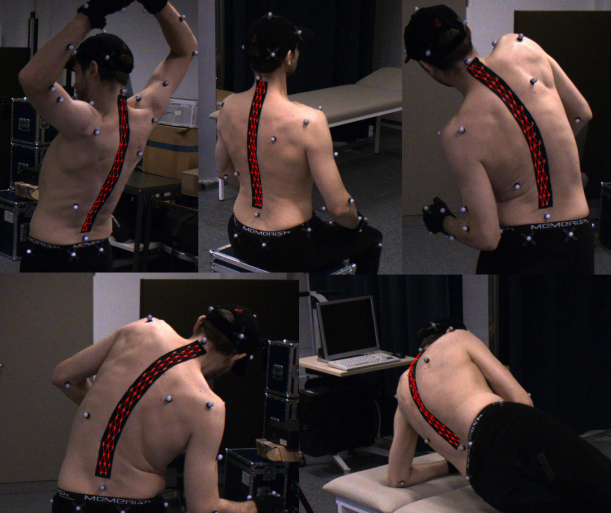}
 \end{center}
 \caption{Tracker results on studio samples.}
 \label{studioresults}
\end{figure}

\textbf{Qualitative results:} In Fig. \ref{occlusionExperiment}, \ref{comparisonToSMPL},\ref{occlusion} and \ref{studioresults} some samples of our tracker on various studio sequences are visualized. Fig. \ref{occlusion} shows an artificial occlusion by a jogging bra, which occludes dots in all views. Even though the Mask R-CNN results are incomplete or include a second faulty detection, the tracking result is precise.

\newcommand{\mywidthhh}{0.24}
\begin{figure}[t]
\begin{center}
 \includegraphics[width=\mywidthhh\linewidth]{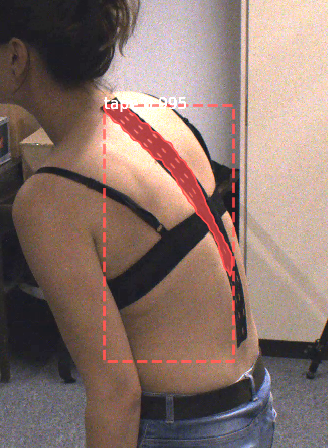}
 \includegraphics[width=\mywidthhh\linewidth]{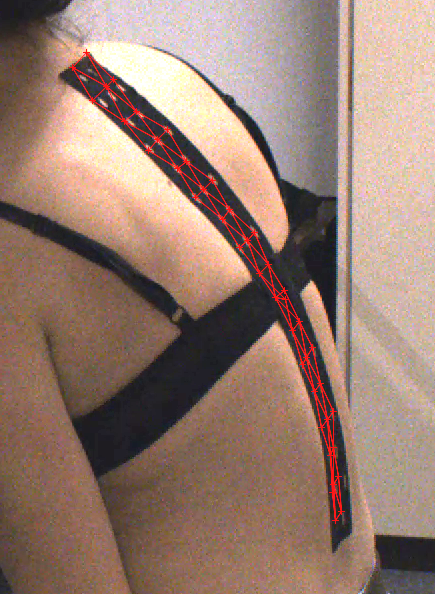} 
 \includegraphics[width=\mywidthhh\linewidth]{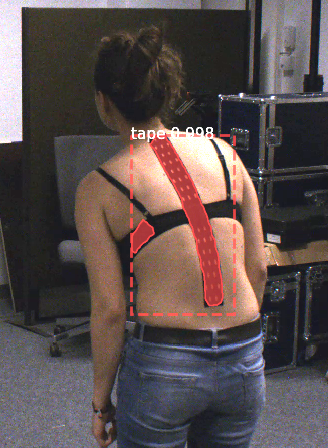}  
 \includegraphics[width=\mywidthhh\linewidth]{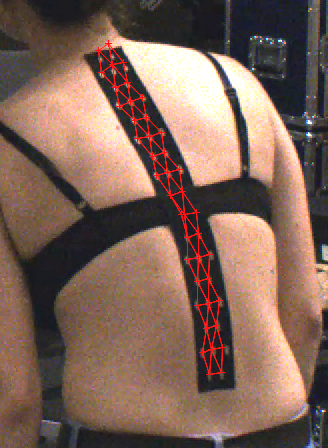}
 \end{center}
 \caption{The proposed spine tracker also works with jogging bras. Even though some dots in the stripe are occluded and the Mask R-CNN detections have a second faulty detection (right side), the overall stripe fitting is accurate.}
 \label{occlusion}
\end{figure}

\textbf{Discussion:} 
The proposed tracking pipeline in Fig. \ref{blockdiagram} is different from many other trackers, as it has no image-based feature encoding and matching, such as the optical flow texture matching in Dynamic FAUST \cite{Bogo:2017} or the image-based connectivity extraction in the mesh suit approach by Hiroaki \etal \cite{Hiroaki:2005}. We claim that since we have only a detector, we can increase the marker density to a higher level. In comparison to Hiroaki \etal \cite{Hiroaki:2005}, who report a marker density of \unit[0.16]{marker/cm$^2$}, using a similar studio setup we achieve a density of \unit[0.22]{marker/cm$^2$}. 

\begin{table}
\begin{center}
\resizebox{\linewidth}{!}{
\centering
\begin{tabular}[h]{c||c|c|c|c|c|c}
    Properties & 4D\cite{Bogo:2017} & IMU\cite{Voinea:2017} & Vicon\cite{Vicon} & OF\cite{Kam:2017} & MS\cite{Hiroaki:2005} & \textbf{ours} \\
    \hline\hline
    model preparations & $(--)$ & $(o)$ & $(-)$ & $(o)$ & $(++)$ & $(++)$\\
    \hline
    precision & $(++)$ & $(o)$ & $(+)$ & $(--)$ & $(+)$ & $(+)$\\
    \hline
    comfort & $(++)$ & $(-)$  & $(-)$ & $(-)$ & $(++)$ & $(++)$  \\
    \hline
    occlusion handling & $(++)$ & $(++)$ & $(--)$ & $(++)$ & $(--)$ & $(+)$  \\
    \hline
    outdoor & $(--)$ & $(++)$ & $(--)$ & $(++)$ & $(-)$ & $(+)$  \\
\end{tabular}
}
\end{center}
\caption{Qualitative comparison of various motion capture systems (estimated by the authors): 4D denotes body painting in Dynamic FAUST \cite{Bogo:2017}, IMU denotes inertial measurement units \cite{Voinea:2017}, OF denotes optical fibers \cite{Kam:2017} and MS denotes a mesh suit \cite{Hiroaki:2005}.} 
\label{comparisonOfMethods}
\end{table}

We would like to conclude this section with a comparison of different spine motion capture systems. In Table \ref{comparisonOfMethods} strengths and weaknesses regarding certain properties are listed. While some comparisons are difficult to assess, the following can be probably agreed upon: Dynamic FAUST is rather precise, and active lightning systems may run into problems if used outdoors. Focussing on our approach we claim that a model preparation of a few seconds is rather short. The precision is in the upper mid-range. The term comfort refers to whether natural movements are possible and includes if a device stands out, falls off, or is heavy. We showed that our framework is capable of occlusion handling with interpolated marker estimation even if multiple markers are occluded in all cameras. Finally, given that cameras come with automatic exposure, the combination with a high contrast marker allows for outdoor use. We want to stress that our tracker uses intra-frame prediction only and no initial pose is given. 

\section{Conclusion}
\label{sec:conclusion}
We introduced a kinesiology tape-based tracking system that is able to capture the human spine in 3D over time. The tape is highly flexible and can follow any spine movements. Several experiments on artificial and studio sequences show that our tracking framework can reliably track markers with high density and is robust in case of occlusions. The method is not limited to the application of spine motion capture and can be adopted to all other sorts of surface tracking. We assume that the tracking performance can be further improved by temporal context, which we intend to investigate in future work.

\section*{Acknowledgements}
\label{sec:acknowledgements}
This work was supported by the Federal Ministry of Education and Research (BMBF), Germany under the project LeibnizKILabor (grant no. 01DD20003) and the AI service center KISSKI (grant no. 01IS22093C), the Center for Digital Innovations (ZDIN) and the Deutsche Forschungsgemeinschaft (DFG) under Germany’s Excellence Strategy within the Cluster of Excellence PhoenixD (EXC 2122). 
The authors also thank Dr. Oliver Müller for fruitful discussions and hints.

{\small
\bibliographystyle{ieee_fullname}
\bibliography{MoCapSOTA/MoCap.bib}

\begin{thebibliography}{10}\itemsep=-1pt

\bibitem{Vicon}
{Vicon Tracker User Guide}.
\newblock \url{https://docs.vicon.com/}, 9 2020.

\bibitem{Bogduk:2005}
Nikolai Bogduk.
\newblock {\em {Clinical anatomy of the lumbar spine and sacrum}}.
\newblock Elsevier Health Sciences, 2005.

\bibitem{Bogo:2016}
Federica Bogo, Angjoo Kanazawa, Christoph Lassner, Peter Gehler, Javier Romero,
  and Michael~J. Black.
\newblock Keep it {SMPL}: Automatic estimation of {3D} human pose and shape
  from a single image.
\newblock In {\em Computer Vision -- ECCV 2016}, Lecture Notes in Computer
  Science, pages 561--578. Springer International Publishing, Oct. 2016.

\bibitem{Bogo:2017}
Federica Bogo, Javier Romero, Gerard Pons-Moll, and Michael~J. Black.
\newblock {Dynamic FAUST: Registering Human Bodies in Motion}.
\newblock In {\em {2017 IEEE Conference on Computer Vision and Pattern
  Recognition (CVPR)}}, pages 5573--5582, 2017.

\bibitem{Campbell:2014}
Amity Campbell, Peter O'Sullivan, Leon Straker, Bruce Elliott, and Machar Reid.
\newblock {Back pain in tennis players: a link with lumbar serve kinematics and
  range of motion}.
\newblock {\em Medicine and science in sports and exercise}, 46(2):351---357,
  February 2014.

\bibitem{Delp:2007}
Scott~L. Delp, Frank~C. Anderson, Allison~S. Arnold, Peter Loan, Ayman Habib,
  Chand~T. John, Eran Guendelman, and Darryl~G. Thelen.
\newblock {OpenSim: Open-Source Software to Create and Analyze Dynamic
  Simulations of Movement.}
\newblock {\em IEEE Trans. Biomed. Eng.}, 54(11):1940--1950, 2007.

\bibitem{DocBra2016}
Alexander Dockhorn, Christian Braune, and Rudolf Kruse.
\newblock Variable density based clustering.
\newblock In {\em 2016 IEEE Symposium Series on Computational Intelligence
  (SSCI)}, pages 1--8, Dec. 2016.

\bibitem{Gurobi:2023}
{Gurobi Optimization, LLC}.
\newblock {Gurobi Optimizer Reference Manual}, 2023.

\bibitem{HacKru2021a}
Hendrik Hachmann, Benjamin Kr{\"u}ger, Bodo Rosenhahn, and Waldo Nogueira.
\newblock Localization of cochlear implant electrodes from cone beam computed
  tomography using particle belief propagation.
\newblock In {\em International Symposium on Biomedical Imaging, ISBI}, april
  2021.

\bibitem{Hajibozorgi:2016}
M. Hajibozorgi and N. Arjmand.
\newblock {Sagittal range of motion of the thoracic spine using inertial
  tracking device and effect of measurement errors on model predictions}.
\newblock {\em Journal of Biomechanics}, 49(6):913--918, 2016.
\newblock SI: Spine Loading and Deformation.

\bibitem{He:2017}
Kaiming He, Georgia Gkioxari, Piotr Doll{\'a}r, and Ross~B. Girshick.
\newblock {Mask {R-CNN}}.
\newblock {\em CoRR}, abs/1703.06870, 2017.

\bibitem{Kam:2017}
Wern Kam, Kieran O{\rq}Sullivan, Mary O{\rq}Keeffe, Sinead O{\rq}Keeffe,
  Waleed~S. Mohammed, and Elfed Lewis.
\newblock {Low cost portable 3-D printed optical fiber sensor for real-time
  monitoring of lower back bending}.
\newblock {\em Sensors and Actuators A-physical}, 265:193--201, 2017.

\bibitem{Lin:2014}
Tsung-Yi Lin, Michael Maire, Serge Belongie, Lubomir Bourdev, Ross Girshick,
  James Hays, Pietro Perona, Deva Ramanan, C.~Lawrence Zitnick, and Piotr
  Doll{\'a}r.
\newblock {Microsoft COCO: Common Objects in Context}, 2014.

\bibitem{Loper:2014}
Matthew Loper, Naureen Mahmood, and Michael~J. Black.
\newblock {MoSh: Motion and Shape Capture from Sparse Markers}.
\newblock {\em ACM Trans. Graph.}, 33(6), nov 2014.

\bibitem{Loper:2015}
Matthew Loper, Naureen Mahmood, Javier Romero, Gerard Pons-Moll, and Michael~J.
  Black.
\newblock {{SMPL}: A Skinned Multi-Person Linear Model}.
\newblock {\em ACM Trans. Graphics (Proc. SIGGRAPH Asia)}, 34(6):248:1--248:16,
  Oct. 2015.

\bibitem{Merriaux:2017}
Pierre Merriaux, Yohan Dupuis, R{\'e}mi Boutteau, Pascal Vasseur, and Xavier
  Savatier.
\newblock {A Study of Vicon System Positioning Performance}.
\newblock {\em Sensors}, 17(7), 2017.

\bibitem{Mueller:2013}
Oliver M{\"u}ller, Michael~Y. Yang, and Bodo Rosenhahn.
\newblock {Slice Sampling Particle Belief Propagation}.
\newblock In {\em {IEEE International Conference on Computer Vision (ICCV)}},
  pages 1129--1136, Dec. 2013.

\bibitem{Pacheco:2015}
Jason Pacheco and Erik Sudderth.
\newblock {Proteins, Particles, and Pseudo-Max-Marginals: A Submodular
  Approach}.
\newblock In Francis Bach and David Blei, editors, {\em {Proceedings of the
  32nd International Conference on Machine Learning}}, volume~37 of {\em
  {Proceedings of Machine Learning Research}}, pages 2200--2208, Lille, France,
  07--09 Jul 2015. PMLR.

\bibitem{Rast:2016}
Fabian~M. Rast, Eveline~S. Graf, André Meichtry, Jan Kool, and Christoph~M.
  Bauer.
\newblock Between-day reliability of three-dimensional motion analysis of the
  trunk: A comparison of marker based protocols.
\newblock {\em Journal of Biomechanics}, 49(5):807--811, 2016.

\bibitem{RosSch2008}
B. Rosenhahn, C. Schmaltz, T. Brox, J. Weickert, D. Cremers, and H.-P. Seidel.
\newblock Markerless motion capture of man-machine interaction.
\newblock In {\em IEEE Conference on Computer Vision and Pattern Recognition},
  2008.

\bibitem{Schepers:2018}
Martin Schepers, Matteo Giuberti, and Giovanni Bellusci.
\newblock {Xsens MVN: Consistent Tracking of Human Motion Using Inertial
  Sensing}.
\newblock {\em White paper}, 03 2018.

\bibitem{Hiroaki:2005}
Hiroaki Tanie, Katsu Yamane, and Yoshihiko Nakamura.
\newblock High marker density motion capture by retroreflective mesh suit.
\newblock In {\em Proceedings of the 2005 {IEEE} International Conference on
  Robotics and Automation, {ICRA} 2005, April 18-22, 2005, Barcelona, Spain},
  pages 2884--2889. {IEEE}, 2005.

\bibitem{Matlab:2019}
Inc. {The MathWorks}.
\newblock {\em {Computer Vision Toolbox}}.
\newblock Natick, Massachusetts, United State, 2019.

\bibitem{Voinea:2017}
Gheorghe-Daniel Voinea, Silviu Butnariu, and Gheorghe Mogan.
\newblock {Measurement and Geometric Modelling of Human Spine Posture for
  Medical Rehabilitation Purposes Using a Wearable Monitoring System Based on
  Inertial Sensors}.
\newblock {\em Sensors}, 17(1), 2017.

\bibitem{Wei:2016}
Shih-En Wei, Varun Ramakrishna, Takeo Kanade, and Yaser Sheikh.
\newblock {Convolutional Pose Machines.}
\newblock {\em CoRR}, abs/1602.00134, 2016.

\end{thebibliography}
}

\end{document}